\documentclass[a4paper,fleqn]{cas-dc}

\usepackage[authoryear]{natbib}
\usepackage{times}
\usepackage{indentfirst}
\usepackage{microtype}      
\usepackage{soul}           
\usepackage{xcolor}         

\usepackage{amsmath}        
\usepackage{amsthm}
\usepackage{amsfonts}       
\usepackage{bbding}         
\usepackage{pifont}         

\usepackage{graphicx}       
\usepackage{booktabs}       
\usepackage{colortbl}       
\usepackage{multirow}       
\usepackage{wrapfig}        
\usepackage{float}          
\usepackage{subcaption}     
\usepackage{overpic}        
\usepackage{rotating}       
\usepackage{ulem}

\usepackage{algorithm}
\usepackage{algorithmic}

\usepackage{url}

\usepackage{listings}       

\usepackage{tikz,pgfplots}  

\usepackage{caption}

\usepackage{nicefrac}       %

\usepackage[capitalize]{cleveref}
\crefname{section}{Sec.}{Secs.}
\Crefname{section}{Section}{Sections}
\Crefname{table}{Table}{Tables}
\crefname{table}{Tab.}{Tabs.}

\def\tsc#1{\csdef{#1}{\textsc{\lowercase{#1}}\xspace}}
\tsc{WGM}
\tsc{QE}

\pgfplotsset{compat=1.18}
\begin{document}
\let\WriteBookmarks\relax
\def\floatpagepagefraction{1}
\def\textpagefraction{.001}

\shorttitle{Improving Anomaly Detection with Foundation-Model Synthesis and Wavelet-Domain Attention}

\shortauthors{Wensheng Wu, et.al.}

\title [mode = title]{Improving Anomaly Detection with Foundation-Model Synthesis and Wavelet-Domain Attention}



\author[1]{Wensheng~Wu}
\credit{Conceptualization, Investigation, Methodology, Experiment, Software, Validation, Visualization, Writing - Original Draft, Writing - Review \& Editing, Data Curation}
\author[1]{Zheming Lu}
\credit{Writing - Original Draft, Writing - Review \& Editing, Data Curation}
\author[2]{Ziqian Lu}
\author[1]{Zewei He}
\author[1]{Xuecheng Sun}
\author[3]{Zhao Wang}
\author[4]{Jungong Han}
\author[5]{Yunlong Yu}
\cormark[1]
\ead{yuyunlong@zju.edu.cn}
\affiliation[1]{
    organization={School of Aeronautics and Astronautics},
    addressline={Zhejiang University}, 
    city={Hangzhou},
    postcode={310027}, 
    country={China}}

\affiliation[2]{
    organization={School of Computer Science and Technology()},
    addressline={}, 
    city={Hangzhou},
    postcode={310018}, 
    country={China}}

\affiliation[3]{
    organization={Ningbo Innovation Center},
    addressline={Zhejiang University}, 
    city={Ningbo},
    country={China}}
    
\affiliation[4]{
    organization={Department of Automation},
    addressline={Tsinghua University}, 
    city={Beijing},
    postcode={100084}, 
    country={China}}   

\affiliation[5]{
    organization={College of Information Science \& Electronic Engineering},
    addressline={Zhejiang University}, 
    city={Hangzhou},
    postcode={310027}, 
    country={China}}

\cortext[1]{Corresponding author}



\begin{abstract}
Industrial anomaly detection faces significant challenges due to the scarcity of anomalous samples and the complexity of real-world anomalies. In this paper, we propose a foundation model-based anomaly synthesis pipeline (FMAS) that generates highly realistic anomalous samples without fine-tuning or class-specific training. Motivated by the distinct frequency-domain characteristics of anomalies, we introduce a Wavelet Domain Attention Module (WDAM), which exploits adaptive sub-band processing to enhance anomaly feature extraction. The combination of FMAS and WDAM significantly improves anomaly detection sensitivity while maintaining computational efficiency. Comprehensive experiments on MVTec AD and VisA datasets demonstrate that WDAM, as a plug-and-play module, achieves substantial performance gains against existing baselines.
\end{abstract}

\let\printorcid\relax
\maketitle

\section{Introduction}
Industrial visual anomaly detection serves as a critical technique for identifying and localizing anomaly in manufactured products, thereby contributing to reductions in anomaly rates and operational labor costs. Due to the inherent scarcity of anomalous samples, recent research has predominantly focused on unsupervised learning paradigms that leverage only normal data during training. These methods can be broadly categorized into two principal groups: 1) embedding-based approaches \citep{simplenet,efficientad,padim,jiang2025localize}, which utilize deep feature representations to detect and localize anomaly; and 2) reconstruction-based approaches \citep{omnial,chen2022utrad,glad}, which attempt to reconstruct the input image and identify anomalies through deviations between the reconstructed and original images.

Despite their utility, relying solely on normal data significantly limits the detection capabilities of these models. To address this limitation, a number of studies have been proposed synthetic anomaly generation strategies. Among these, the non-generative methods \citep{cutpaste,draem,GLASS,wind-turbine} synthesize anomalies by applying deterministic or stochastic
perturbations to normal images.
However, these synthesized anomalies frequently lack visual realism and fail to capture the complex statistical properties of real-world anomaly. In contrast, the generative methods, such as DFMGAN \citep{dfmgan} and RealNet \citep{realnet}, employ generative adversarial networks or diffusion models to produce more photorealistic anomalies. Nevertheless, these methods cannot be deployed immediately in new scenarios, as they rely on training and typically require additional adaptation or fine-tuning.

\begin{figure}
\centering
\includegraphics[width=\columnwidth]{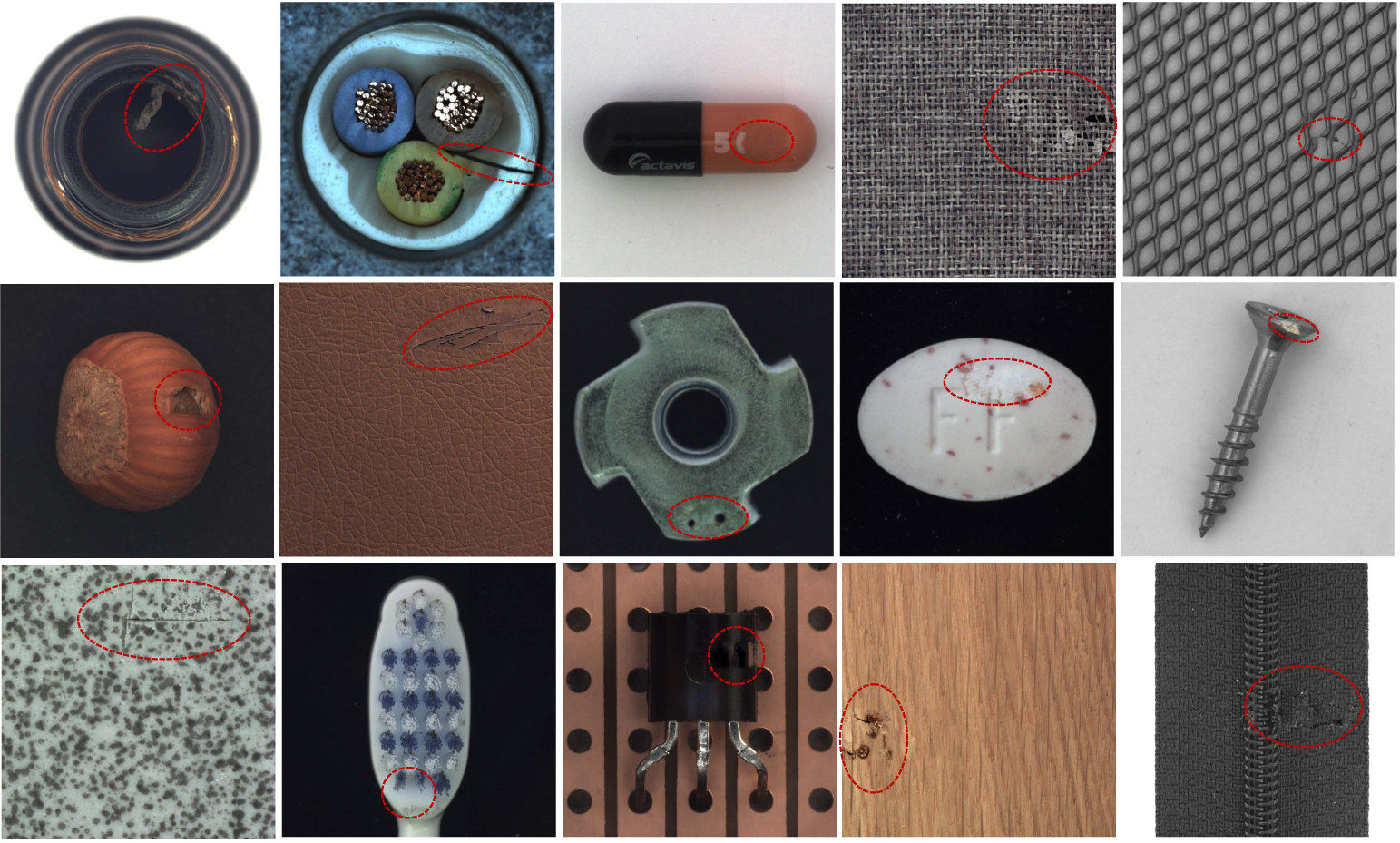}
\caption{Qualitative results showing anomaly samples generated by the proposed FMAS on the MVTec AD dataset, with red contours highlighting the anomalous regions.}
\label{fig:generated-samples}
\end{figure}

In this paper, we propose a Foundation Model-based Anomaly Synthesis (FMAS) pipeline, which integrates the capabilities of foundation models, including GPT-4 \citep{gpt}, Segment Anything Model (SAM) \citep{sam}, and Stable Diffusion \citep{sd}, to synthesize highly realistic anomalous data without the need for model fine-tuning or class-specific training. Specifically, GPT-4 is employed to automatically generate descriptive prompts that guide the image synthesis process. SAM is used to segment foreground objects and provide spatial context for subsequent anomaly generation. Stable Diffusion is then utilized to inpaint plausible anomalies into the images based on the generated prompts. However, due to the lack of task-specific constraints, the generation process may involve a certain degree of randomness, which can occasionally result in trivial or excessively distorted samples that do not meet the desired quality standards. To address this issue, a filtering mechanism, referred to as the Selector, is introduced to automatically exclude such low-quality outputs. Representative qualitative results are presented in \cref{fig:generated-samples}.

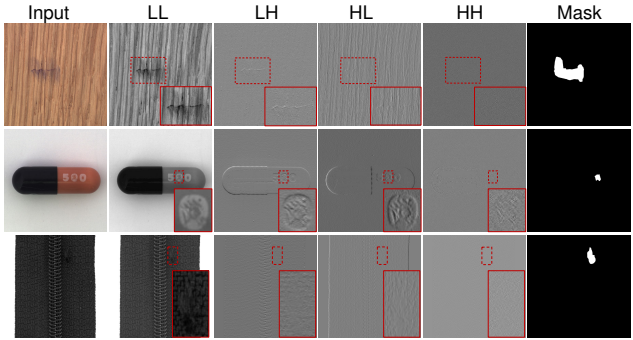
\begin{figure}
    \centering
    \begin{overpic}[width=0.99\linewidth]{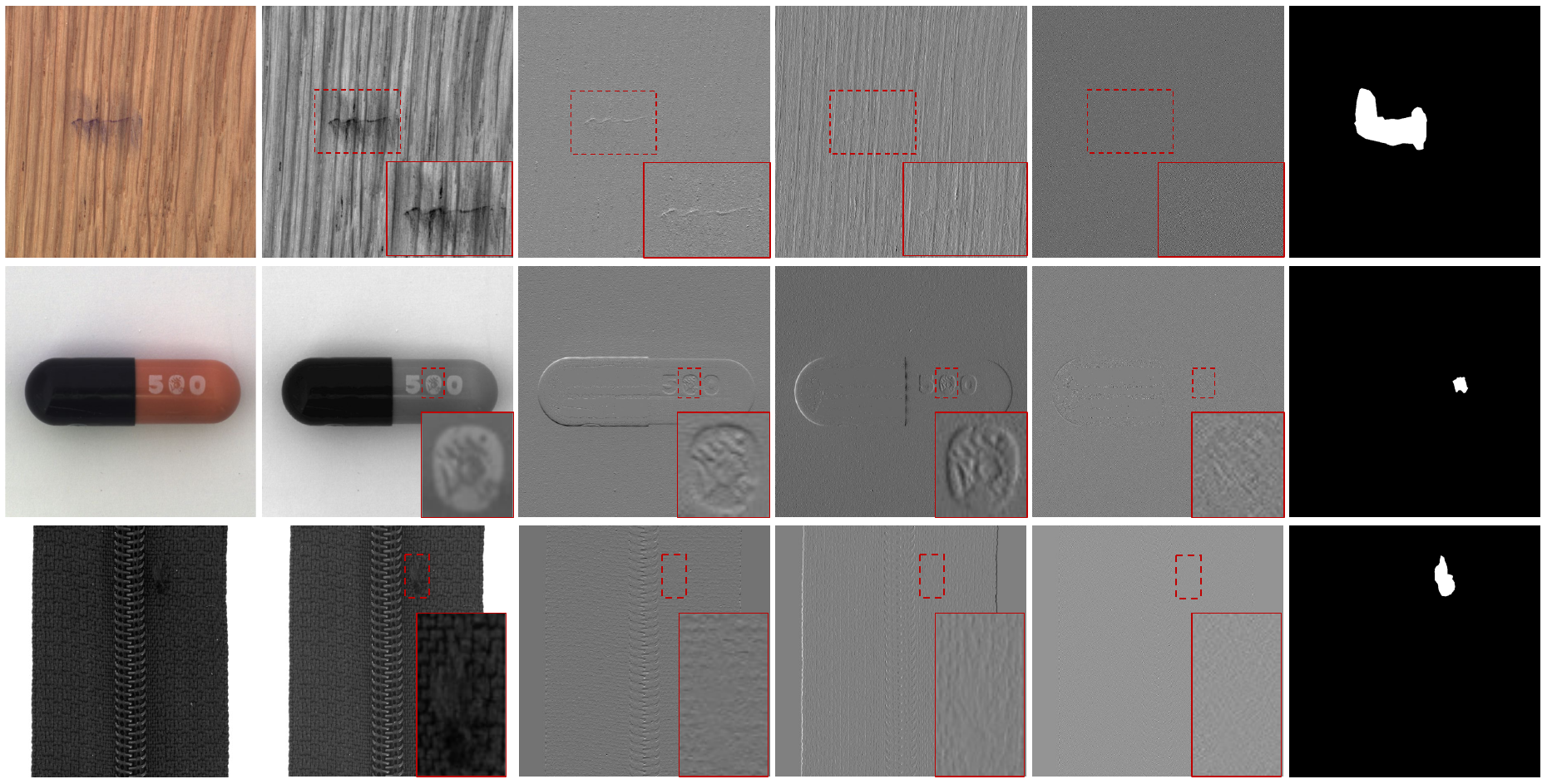}
    \put(4,51){\scriptsize Input}
    \put(23,51){\scriptsize LL}
    \put(40,51){\scriptsize LH}
    \put(55,51){\scriptsize HL}
    \put(72,51){\scriptsize HH}
    \put(88,51){\scriptsize Mask}
    \end{overpic}
    \caption{Visualization of anomaly saliency in the four wavelet sub-bands. The leftmost column presents the original images, the middle columns show the LL, LH, HL, and HH sub-bands, and the rightmost column illustrates the corresponding masks. Red dashed boxes indicate defect locations, and red solid boxes highlight magnified regions. All samples are drawn from the MVTec AD dataset.}
    \label{fig:wavelet-results}
\end{figure}

We use the discrete wavelet transform (DWT) to analyze both synthetic and real anomalies, capturing their spatial-frequency characteristics. This analysis reveals that anomalous features exhibit varying saliency across the four sub-bands (LL, LH, HL, HH), as illustrated in \cref{fig:wavelet-results}, with defect locations highlighted and magnified for clarity. To further quantify this observation, \cref{tab:wavelet_saliency} reports pixel-level variation measurements, computed by accumulating absolute differences between normal and anomalous image pairs within each class, averaging across the dataset, and normalizing across sub-bands. The results indicate that anomalous characteristics are unevenly distributed across sub-bands, highlighting the potential benefit of sub-band-aware modeling for effective anomaly detection.

Building on these findings, we propose separate processing of image components based on their distinct frequency characteristics. Specifically, our proposed Wavelet Domain Attention Module (WDAM) dynamically assigns learnable weights to each wavelet sub-band, where the weight distribution is optimized to amplify anomaly-sensitive frequency components while suppressing irrelevant features. WDAM first decomposes input features into wavelet sub-bands, then applies adaptive attention weights to each sub-band based on anomaly saliency. The weighted features are reconstructed via inverse discrete wavelet transform, effectively amplifying anomaly-related patterns while preserving spatial information. By dynamically enhancing frequency-specific features, WDAM significantly improves anomaly sensitivity and feature discriminability in anomaly detection tasks. WDAM operates as a plug-and-play module that can be seamlessly integrated into existing network architectures.

In summary, the contributions are summarized as follows.

\begin{itemize}
    \item We introduce a Foundation Model-based Anomaly Synthesis (FMAS) pipeline that generates highly realistic anomalous samples \textbf{without requiring model fine-tuning or class-specific training}. With this pipeline, we construct two synthetic datasets perfectly aligned with the MVTec AD and VisA benchmarks.
    
    \item Based on the frequency-domain analysis, we introduce the Wavelet Domain Attention Module (WDAM), a \textbf{plug-and-play} component that operates at the wavelet sub-band level. By adaptively learning the importance of each frequency band, WDAM significantly enhances anomaly detection through feature enhancement.

    \item Experiments on MVTec AD \citep{bergmann2019mvtec} and VisA \citep{zou2022spot} demonstrate that the combination of FMAS and WDAM yields consistent improvements in anomaly detection performance. Further evaluation on two additional network architectures confirms that WDAM enhances anomaly feature extraction and exhibits broad applicability.
    
\end{itemize}

\begin{table*}
\caption{Quantitative comparison of anomaly prominence across different wavelet sub-bands on our synthesized dataset based on the MVTec AD dataset. The reported values represent the relative degree of pixel-level change within each sub-band after anomaly synthesis, with higher values indicating greater relative variation.}\label{tab:wavelet_saliency}
\resizebox{\linewidth}{!}{
\begin{tabular}{lccccc ccccc ccccc}
\toprule
\textbf{Class} & bottle & cable & capsule  & carpet & grid   & hazelnut & leather & metal nut & pill & screw   & tile & toothbrush & transistor & wood  & zipper\\
\midrule
 \textbf{LL}  & 0.69  & 0.84 & 0.90 & 0.72 & 0.76 & 0.76 & 0.69 & 0.78 & 0.81 & 0.75 & 0.83 & 0.80 & 0.81 & 0.70 & 0.70 \\
\textbf{LH}  & 0.10 & 0.07 & 0.05 & 0.12 & 0.10 & 0.08 & 0.14 & 0.10 & 0.08 & 0.10  & 0.08 & 0.08 & 0.07 & 0.09 & 0.12 \\
 \textbf{HL}  & 0.10 & 0.07 & 0.04 & 0.13& 0.10 & 0.08 & 0.13 & 0.09  & 0.08 & 0.10 & 0.08 & 0.10 & 0.08 & 0.17 & 0.14 \\
\textbf{HH} & 0.10 & 0.02 & 0.02 & 0.03 & 0.03 & 0.08 & 0.04 & 0.03 & 0.03 & 0.04 & 0.02 & 0.03 & 0.04 & 0.04 & 0.04 \\
\bottomrule
\end{tabular}
}
\end{table*}

The rest of this paper is organized as follows. \cref{sec:relatedwork} reviews the related work. \cref{sec:method} details the proposed anomaly detection method. \cref{sec:experiment} provides extensive experiments and benchmark evaluations, followed by the conclusion in \cref{sec:conclusion}.

\section{Related work} \label{sec:relatedwork}

\subsection{Anomaly synthesis method}

In unsupervised anomaly detection, most methods are trained solely on normal samples due to the scarcity or absence of real defective data. To compensate for this limitation, a wide range of anomaly synthesis techniques have been proposed. These methods can be broadly classified into two categories: non-generative methods that apply hand-crafted or rule-based transformations to simulate anomalies, and generative methods that rely on deep generative models to produce semantically coherent and visually realistic defect patterns. 

Non-generative methods \citep{cutpaste,nsa,draem,memseg,cdo,simplenet}, construct synthetic anomalies through deterministic or stochastic perturbations applied directly to normal images or their feature space, often without requiring model training. Among these, CutPaste~\citep{cutpaste} simulates anomalies via two strategies: Scar, which removes a narrow region and fills it with random colors, and CutPaste, which pastes large patches within the same image to create spatial disruptions. NSA~\citep{nsa} improves upon this by applying Poisson image editing for smoother boundaries, object mask segmentation, and patch resizing to enhance realism. Both methods derive anomalies from within the original image. In contrast, DRAEM~\citep{draem} introduces external textures from the DTD dataset and blends them using Perlin noise masks. MemSeg~\citep{memseg} further extends this approach with foreground masks and additional synthetic strategies involving geometric and color augmentations followed by patch reordering. CDO~\citep{cdo} diverges by sampling Gaussian noise to replace multiple regions in the image, requiring no auxiliary data. Unlike the above, SimpleNet~\citep{simplenet} operates in the feature space by injecting Gaussian noise into latent features to simulate anomalies. While these approaches are diverse and effective, they often rely on handcrafted rules or limited sources of variability, which constrain the visual realism of the synthesized anomalies and thus motivate the development of generative strategies capable of producing more realistic and semantically diverse anomalies.

Generative methods, by contrast, aim to synthesize anomalies using data-driven generative models, enabling greater control over visual quality and semantic diversity, such as \citep{sdgan,defect-gan,dfmgan,realnet,anomalydiffusion}. SDGAN~\citep{sdgan} is designed with a dual-generator and four-discriminator architecture, and has demonstrated the capability to synthesize visually convincing defects on commutator cylinder surface images. Defect-GAN~\citep{defect-gan} employs an encoder–decoder framework that generates defect samples by simulating a damage-and-repair cycle. Despite their effectiveness, both approaches operate at the full-image level, which may distort normal regions and degrade background fidelity. DFMGAN~\citep{dfmgan} introduces a mask generation module into the StyleGAN2 backbone, thus localizing the synthesis process to predefined anomalous regions and preserving the surrounding normal content. RealNet~\citep{realnet} replaces GAN-based generators with a denoising diffusion probabilistic model (DDPM), which exhibits greater training stability and produces finer texture details. AnomalyDiffusion~\citep{anomalydiffusion} further advances this line of research by leveraging a pre-trained Latent Diffusion Model (LDM) to perform few-shot anomaly generation. Nevertheless, it still requires limited task-specific training to adapt the model to the target domain.

In contrast, our pipeline leverages the strong priors and generalization capabilities of foundation models to synthesize high-fidelity anomaly samples without any training.

\subsection{Wavelet transform application in image processing}
Wavelet transform has emerged as a robust and versatile analytical tool in signal and image processing, owing to its capacity for joint spatial-frequency localization. This property allows for the effective decomposition of signals into multiple scales, making wavelet-based approaches particularly well suited for tasks that involve texture, edge, and structural feature analysis. Consequently, wavelet transform has been widely adopted in a range of image processing applications, such as image compression \citep{da-dwt}, restoration \citep{restoration}, edge detection \citep{edge-detection}, denoising \citep{denoising}, segmentation \citep{segmentation} \citep{bwg,WTConv} and classification \citep{WTConv}, among others.

\subsection{Attention mechanism}

Attention mechanism \citep{attention-survey} is inspired by the human visual system and enables dynamic weighting of the input, emphasizing task-relevant information while suppressing irrelevant parts. This mechanism has been realized through diverse architectural designs, such as channel attention \citep{senet}, a combination of channel attention and spatial attention \citep{CBAM}, or self-attention\citep{vit}. 

Many studies \citep{wa-cnn,dwan,stwanet,swa,EawT} have explored the integration of wavelet transforms with attention mechanisms. WA-CNN~\citep{wa-cnn} applies a wavelet transform to the input, discards the HH sub-band, and treats the remaining sub-bands as query, key and value to perform attention computation. STWANet~\citep{stwanet} concatenates the sub-bands obtained from the wavelet transform to form a new feature representation. SWA~\citep{swa} concatenates the wavelet sub-bands along the spatial dimension and applies spatial attention to the resulting features. DWAN~\citep{dwan} applies both channel and spatial attention to the wavelet domain features, and then uses the resulting attention weights to refine the original spatial domain features. In contrast, EawT~\citep{EawT} performs the allocation of both channel and spatial attention entirely in the wavelet domain before transforming the features back into the spatial domain.

While existing methods typically apply conventional attention mechanisms uniformly across wavelet-domain features, our approach introduces a paradigm shift by treating individual wavelet sub-bands as semantically distinct components and modeling them accordingly.

\section{Method}
\label{sec:method}

\begin{figure}
\centering
\includegraphics[width=\columnwidth]{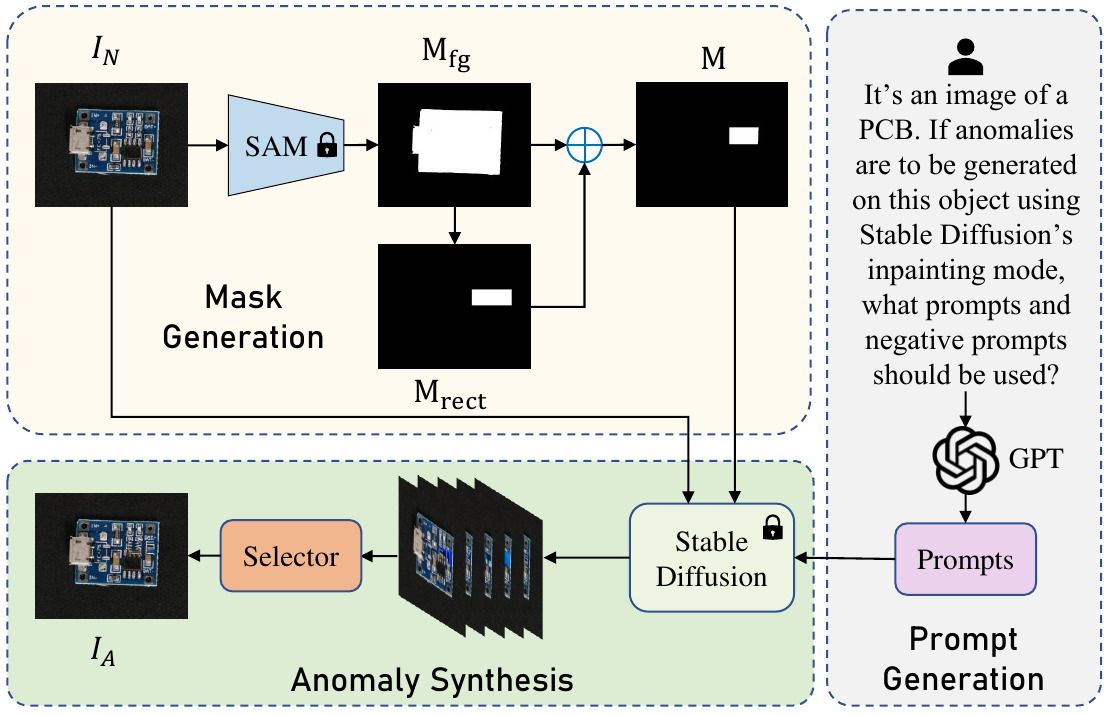}
\caption{Illustration of the proposed foundation model-based anomaly
synthesis pipeline (FMAS).}
\label{fig:Anomaly Synthesis pipeline}
\end{figure}

In this section, we present our anomaly detection pipeline. First, we propose a training-free anomaly synthesis method to generate training data for the detection model. Next, we introduce the Wavelet Domain Attention Module (WDAM), a key component that enables the model to learn discriminative representations from different frequency distributions. Finally, we demonstrate that WDAM serves as a plug-and-play module, seamlessly enhancing existing anomaly detection methods.


\subsection{Anomaly synthesis based on foundation models}

\begin{figure*}[!t]
\centering
\includegraphics[width=\textwidth]{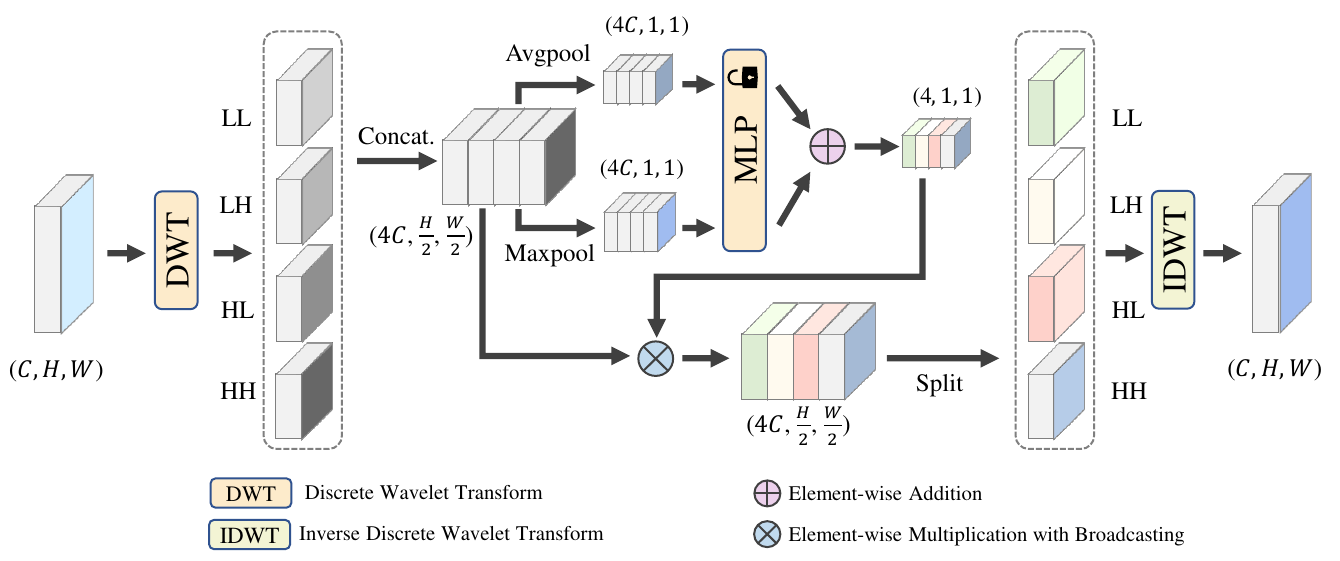}
\caption{The proposed Wavelet Domain Attention Module (WDAM) processes input feature maps through a frequency-aware attention mechanism. First, WDAM decomposes the spatial features into multiple frequency sub-bands using Discrete Wavelet Transform (DWT). An adaptive attention mechanism then selectively enhances or suppresses features in each sub-band based on their relevance to anomaly detection. Finally, the refined frequency components are reconstructed back into the spatial domain via Inverse Discrete Wavelet Transform (IDWT), effectively preserving critical defect-related patterns while suppressing noise. }
\label{fig:WDAM}
\end{figure*}

As shown in \cref{fig:Anomaly Synthesis pipeline}, we propose Foundation Model-based Anomaly Synthesis (FMAS), which integrates GPT, SAM, and Stable Diffusion to generate structurally consistent and semantically plausible anomalies without additional training. In this pipeline, Stable Diffusion operates on three inputs: a normal (anomaly-free) image $I_N$, a binary mask $M$, and text prompts. The mask $M$ is constructed through our customized strategy in which SAM provides the foreground object mask, while the prompts are generated by GPT to supply semantic guidance.

The construction of the mask $M$ is central to the synthesis process. Prior methods such as DRAEM \citep{draem} and RealNet \citep{realnet} rely on Perlin noise masks, but these lack shape controllability and often produce fragmented regions, which hinders the synthesis of high-quality anomalies. To address this, we introduce a rectangular mask generation strategy tailored to the diffusion process. The final mask is obtained as the intersection of a foreground mask and a randomly generated rectangular mask:
\begin{equation}
\label{eq:M}
M = M_{fg} \cap M_{rect},
\end{equation}
where $M_{fg}$ and $M_{rect}$ denote the foreground and rectangular masks, respectively. The foreground mask corresponds to the principal object in $I_N$ and is extracted using the Segment Anything Model (SAM) \citep{sam}, which provides robust segmentation given simple prompts. A point is then randomly sampled within the foreground mask to define the center of $M_{rect}$, whose area is constrained relative to the object as follows:
\begin{equation}
\label{eq:SMrect}
\sum\limits_{i=1}^{h} \sum\limits_{j=1}^{w} M_{rect}^{i,j} = \alpha \sum\limits_{i=1}^{h} \sum\limits_{j=1}^{w} M_{fg}^{i,j},
\end{equation}
where $\alpha$ is set to 0.1 and $(i,j)$ denotes spatial positions in the binary mask. The aspect ratio of the rectangle is randomly sampled within a predefined range.

With the mask established, we then employ a large language model (LLM) to generate context-aware prompts for Stable Diffusion’s inpainting process. The LLM is instructed with the class label of the object (e.g., “It’s an image of a [class label]. If anomalies are to be generated on this object using Stable Diffusion’s inpainting mode, what prompts and negative prompts should be used?”). This step leverages the rich prior knowledge of the LLM to produce both descriptive and negative prompts, guiding the inpainting toward semantically meaningful anomalies.

Finally, given $I_N$, the generated prompts, and mask $M$, Stable Diffusion synthesizes anomalous content within the specified region. For each image-mask pair, five variants are generated by varying the sampling seed while keeping other parameters fixed. To control anomaly intensity, we introduce an LPIPS-based filtering mechanism. For each variant $I_A^{(k)}$, its LPIPS distance to $I_N$ is computed, and the candidate closest to a reference threshold $\tau$ is selected:
\begin{equation}
\label{eq:lpips_select}
\hat{I}A = \arg\min{I_A^{(k)}} | \text{LPIPS}(I_A^{(k)}, I_N) - \tau |,
\end{equation}
where $\hat{I}_A$ denotes the final anomaly sample. This filtering avoids both trivial and overly distorted results, ensuring realistic yet perceptually distinct anomalies.

\begin{table}[!t]
\centering
\caption{Quantitative results of anomaly image realism between our method and the alternative anomaly synthesis approaches on the MVTec-AD dataset.}
\label{tab:synthesis}
\resizebox{0.99\linewidth}{!}{
\begin{tabular}{p{1.5cm} p{2cm} p{2.5cm} p{1.5cm}}
\toprule
Metric & GLASS \citep{GLASS} & RealNet \citep{realnet} & \textbf{Ours} \\
\midrule
FID~$\downarrow$ & 177.51 & 69.64 & 56.99 \\
LPIPS~$\downarrow$ & 0.280 & 0.147 & 0.130 \\
\bottomrule
\end{tabular}
}
\end{table}

To this end, we construct a dataset where each anomalous image corresponds one-to-one with an original normal image from the training set. As demonstrated in \cref{tab:synthesis}, our method generates significantly more realistic anomalies compared to existing synthesis approaches \citep{GLASS,realnet}, as evidenced by lower Fréchet Inception Distance (FID) \citep{FID} and Learned Perceptual Image Patch Similarity (LPIPS) \citep{LPIPS} scores. FID measures the distributional distance between generated and real images in the feature space of a pretrained Inception network, while LPIPS assesses perceptual similarity based on deep feature embeddings. For both metrics, lower values indicate higher similarity to real images, reflecting improved realism in the synthesized anomalies.

\subsection{Wavelet domain attention module}
In this section, we present the Wavelet Domain Attention Module (WDAM). As shown in \cref{fig:WDAM}, WDAM first decomposes the input features into four frequency sub-bands via discrete wavelet transform. It then dynamically assigns learnable attention weights to each sub-band, enhancing discriminative feature representation. Finally, the weighted sub-bands are fused through inverse discrete wavelet transform to reconstruct the refined features.

We follow \citep{WTConv} and  implement both the discrete wavelet transform (DWT) and inverse DWT (IDWT) using fixed convolutional kernels. This design ensures higher computational efficiency and enables seamless integration of discrete wavelet transforms into neural networks. Specifically, the wavelet bases are encoded as a set of non-trainable convolutional filters applied to the input feature maps to generate the four sub-bands: LL, LH, HL, and HH as follows:
\begin{equation} \label{eq:wtconv}
f_{LL},f_{LH},f_{HL},f_{HH} = W(\theta,f),
\end{equation}
where $f$ denotes the input feature maps, $\theta$ denotes convolutional kernel parameters assigned according to the wavelet basis, $W(\cdot)$ denotes the wavelet convolution, $f_{LL},f_{LH},f_{HL},f_{HH}$ denotes the four wavelet sub-bands obtained from the input. The inverse transform is similarly realized using transposed convolutions with corresponding reconstruction filters as follows:
\begin{equation}
\label{eq:wtconv}
\tilde{f} = W'(\tilde{\theta},(f_{LL},f_{LH},f_{HL},f_{HH})),
\end{equation}
where $\tilde{\theta}$ denotes kernel parameters of transposed convolutions, $W'(\cdot)$ denotes the inverse wavelet convolution, $\tilde{f}$ denotes the reconstructed feature. 




Given an input feature map \( \mathbf{X} \in \mathbb{R}^{C \times H \times W} \), we first apply a discrete wavelet transform to decompose it into four frequency sub-bands: LL, LH, HL, and HH, each with dimensions \( \mathbb{R}^{C \times \frac{H}{2} \times \frac{W}{2}} \). Specifically, the LL corresponds to the low-frequency sub-band, whereas LH, HL, and HH represent the high-frequency sub-bands. All sub-bands are then concatenated along the channel dimension to form a joint wavelet representation \( \mathbf{X}_w \in \mathbb{R}^{4C \times \frac{H}{2} \times \frac{W}{2}} \).

In order to enhance feature aggregation and retain richer information, we follow \citep{CBAM} and apply both max pooling and average pooling to the feature maps. The subsequent computation on \( \mathbf{X}_w \) is performed as follows:
\begin{equation}
\label{eq:pooling}
\begin{aligned}
    &\mathbf{X}_w^{avg} = \text{AvgPool}(\mathbf{X}_w),\\
    &\mathbf{X}_w^{max} = \text{MaxPool}(\mathbf{X}_w),
\end{aligned}
\end{equation}
where $\mathbf{X}_w^{avg}$ and $\mathbf{X}_w^{max}$ denote outputs of average pooling and max pooling operations, respectively, both producing spatially compressed features of size \( \mathbb{R}^{4C \times 1 \times 1} \). 

We then process the pooled features $\mathbf{X}_w^{avg}$ and $\mathbf{X}_w^{max}$ through a trainable multilayer perceptron (MLP), and combine their outputs via element-wise summation to generate the wavelet domain attention weights:
\begin{equation}
\label{eq:pooling}
\begin{split}
\mathbf{A}_w = f_{MLP}(\mathbf{X}_w^{avg}) + f_{MLP}(\mathbf{X}_w^{max}),
\end{split}
\end{equation}
where $f_{MLP}$ denotes the MLP in the network, \(\mathbf{A}_w \in \mathbb{R}^{4 \times {1} \times {1}} \) denotes the attention weights for the four wavelet sub-bands.

\begin{figure}
\centering
\includegraphics[width=\columnwidth]{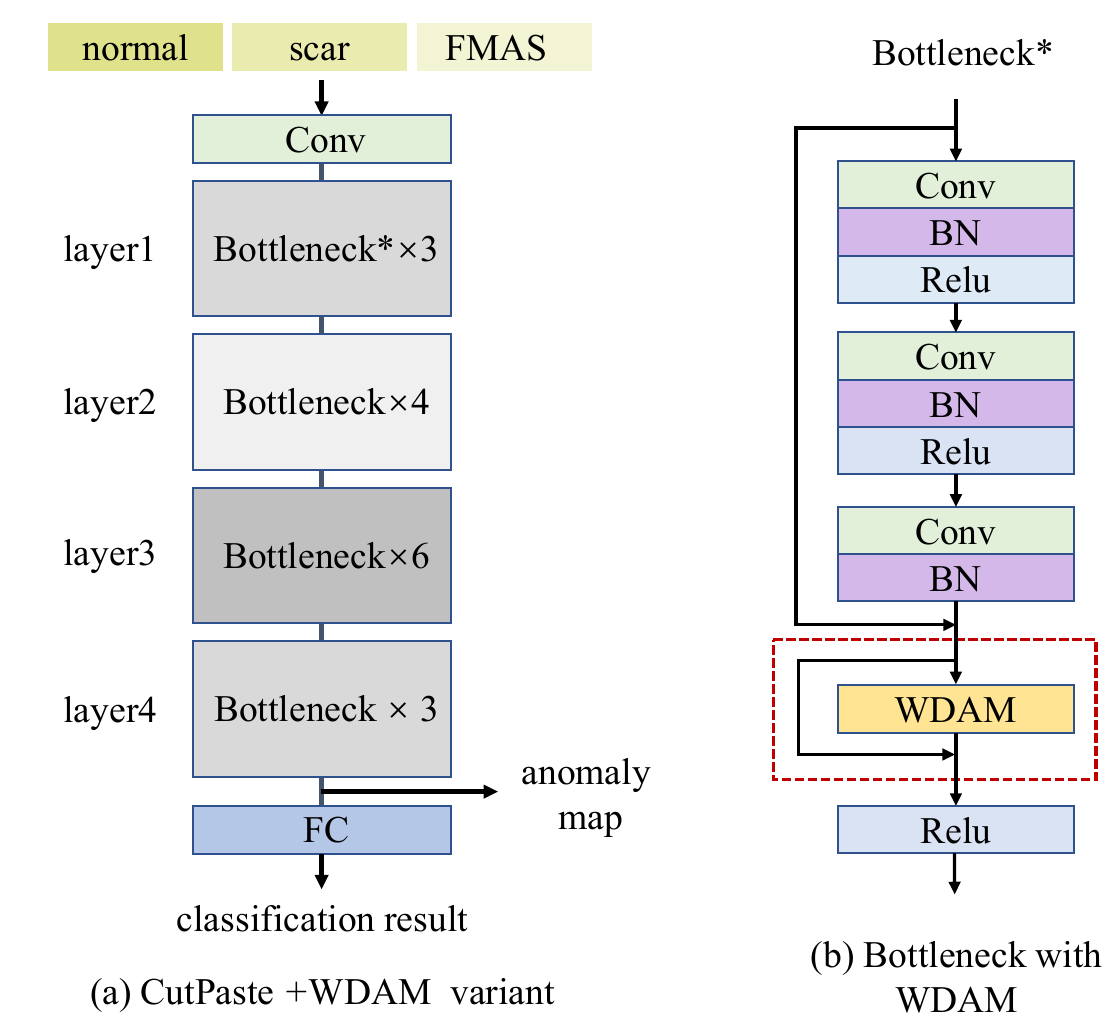}
\caption{(a) The architecture of CutPaste with the proposed WDAM, wherein Bottleneck* denotes a modified bottleneck block incorporating WDAM, as elaborated in (b). The elements enclosed within the red dashed box indicate the additional components introduced in comparison to the original bottleneck configuration.}
\label{fig:cutpaste_with_wdam}
\end{figure}

The attention vector $\mathbf{A}_w$ is broadcast to $\tilde{\mathbf{A}}_w \in \mathbb{R}^{4C \times \frac{H}{2} \times \frac{W}{2}}$ and applied to $\mathbf{X}_w$ via element-wise multiplication, enabling adaptive scaling as follows:
\begin{equation}
\label{eq:pooling}
\begin{split}
\tilde{\mathbf{X}}_w = \mathbf{X}_w \odot \tilde{\mathbf{A}}_w,
\end{split}
\end{equation}
where $\odot$ denotes element-wise multiplication, $\tilde{\mathbf{X}}_w$ denotes the attention-modulated feature. The modulated feature \(\tilde{\mathbf{X}}_w \) is then split back into four sub-bands, which are fed into an Inverse DWT (IDWT) to reconstruct the spatial-domain output feature map \( \tilde{\mathbf{X}} \in \mathbb{R}^{C \times H \times W} \).

Subsequently, the features are transformed back to the spatial domain through inverse wavelet convolution. It is important to highlight that the attention mechanism applied in the wavelet domain effectively enhances the feature representations while the spatial dimensions of the features remain unchanged throughout this process. 


\subsection{Apply WDAM to existing baselines}

Our WDAM operates as a plug-and-play component that seamlessly integrates with existing architectures. As illustrated in \cref{fig:cutpaste_with_wdam} using CutPaste \citep{cutpaste} as an example, WDAM is inserted into each bottleneck in layer1, where feature processing splits into two parallel pathways: (1) a WDAM-enhanced branch and (2) an identity bypass. The outputs are combined through residual summation and ReLU activation, formulated as:
\begin{equation}
\label{eq:pooling}
\begin{split}
\mathbf{X}' = \text{ReLU}(\mathbf{X} + {\text{F}}(\mathbf{X})),
\end{split}
\end{equation}
where $\mathbf{X}$ denotes the output of the previous blocks, $\text{F}(\cdot)$ denotes the WDAM. This design achieves three key advantages: (1) enhanced feature discriminability through wavelet-domain attention, (2) preserved network stability via residual learning, and (3) architecture-agnostic integration without structural modifications.

\begin{table*}[t]
\centering
\caption{Evaluation results of CutPaste and our variants on MVTec AD dataset, with image-level AUROC, pixel-level AUROC, and PRO metrics. FM is short for foreground mask. \textbf{Bold} and \underline{underlined} indicate the best and second-best results, respectively.}
\label{tab:cutpaste-mvtec}
\renewcommand{\arraystretch}{1.2} 
\setlength{\tabcolsep}{3pt} 
\begin{tabular}{>{\raggedright\arraybackslash}p{2.5cm}|%
                >{\centering\arraybackslash}p{3.3cm}%
                >{\centering\arraybackslash}p{3.3cm}%
                >{\centering\arraybackslash}p{3.3cm}%
                >{\centering\arraybackslash}p{3.3cm}}
\toprule
\textbf{Category} & \textbf{CutPaste} & \textbf{+FMAS} & \textbf{+FM} & \textbf{+WDAM (Final Model)}  \\
\midrule
Bottle & (\textbf{100.0}, 68.06, 39.84) & (\textbf{100.0}, 81.18, 66.07) & (\textbf{100.0}, \underline{83.73}, \textbf{71.35}) & (\textbf{100.0}, \textbf{86.82}, \underline{71.29})  \\
Cable & (88.44, 80.39, 43.13) & (\underline{94.10}, \underline{87.80}, 59.77) & (91.55, 85.48, \underline{62.97}) & (\textbf{97.91}, \textbf{90.79}, \textbf{74.14}) \\
Capsule & (\textbf{100.0}, 89.67, 65.91) & (96.38, \underline{92.58}, 74.94) & (96.66, 91.04, \underline{75.15}) & (\underline{97.50}, \textbf{93.26}, \textbf{77.39})  \\
Carpet & (81.77, 88.02, 23.36) & (\underline{92.12}, \underline{95.97}, \underline{68.20}) & (91.59, 95.59, 62.93) & (\textbf{98.66}, \textbf{96.70}, \textbf{81.25})  \\
Grid & (90.64, 77.65, 56.42) & (99.14, \textbf{92.08}, \textbf{68.69}) & (\underline{99.16}, \underline{90.69}, 63.96) & (\textbf{99.89}, 88.78, \underline{66.29}) \\
Hazelnut & (98.58, 84.77, 69.51) & (\underline{99.48}, \underline{92.31}, \underline{82.12}) & (98.89, 91.40, 73.78) & (\textbf{99.73}, \textbf{92.54}, \textbf{84.02})\\
Leather & (\textbf{100.0}, \textbf{96.76}, 83.52) & (\textbf{100.0}, 96.03, \textbf{93.31}) & (\textbf{100.0}, \underline{96.53}, \underline{92.89}) & (\textbf{100.0}, 94.63, 89.29)  \\
Metal nut & (93.16, 61.40, 44.25) & (\textbf{98.21}, \underline{78.31}, \underline{62.58}) & (\textbf{98.21}, 76.17, 56.31)  & (\underline{98.00}, \textbf{84.36}, \textbf{65.51})\\
Pill & (89.37, \underline{82.72}, 59.04) & (\underline{94.96}, 71.17, \underline{70.57}) & (\textbf{95.23}, 76.24, \textbf{74.44}) & (94.72, \textbf{85.09}, 65.06)  \\
Screw & (75.84, 81.13, 45.31) & (\underline{90.59}, 88.31, 57.11) & (87.03, \textbf{91.96}, \textbf{67.74}) & (\textbf{93.48}, \underline{90.71}, \underline{66.23})  \\
Tile & (97.21, 87.11, 70.96) & (98.18, 90.69, \underline{73.49}) & (\underline{99.33}, \underline{92.56}, 67.85) & (\textbf{100.0}, \textbf{93.92}, \textbf{79.79})  \\
Toothbrush & (\textbf{98.61}, 89.19, 57.22) & (89.07, 89.84, 49.80) & (\underline{95.83}, \underline{90.18}, \textbf{68.01}) & (\underline{95.83}, \textbf{90.27}, \underline{60.45}) \\
Transistor & (95.60, \underline{69.06}, 45.17) & (\underline{98.61}, \textbf{71.52}, \underline{59.60}) & (98.06, 65.53, 36.48) & (\textbf{98.88}, 68.03, \textbf{62.55})\\
Wood & (\underline{99.44}, 83.37, \textbf{69.97}) & (97.95, 78.01, 58.42) & (97.84, \textbf{84.33}, \underline{67.35}) & (\textbf{99.47}, \underline{83.70}, 60.92) \\
Zipper & (\textbf{99.96}, \textbf{95.78}, \textbf{89.06}) & (99.05, 91.99, 76.69) & (98.69, 91.74, 74.19) & (\underline{99.95}, \underline{94.21}, \underline{83.74})  \\
\midrule
\textbf{Average} & (93.23, 82.34, 57.51) & (96.52, 86.52, \underline{68.09}) & (\underline{96.54}, \underline{86.88}, 67.69) & (\textbf{98.00}, \textbf{88.28}, \textbf{72.04}) \\
\bottomrule
\end{tabular}
\end{table*}

\section{Experiment}
\label{sec:experiment}

\subsection{Datasets and metrics}
\textbf{Datasets.} The experimental evaluation was conducted primarily on two popular benchmarks, i.e., MVTec AD \citep{bergmann2019mvtec} and VisA \citep{zou2022spot}. The MVTec AD dataset comprises 15 distinct categories, including 5 texture-based classes and 10 object-based classes. The training set includes a total of 3,629 anomaly-free images. The test set provides normal and anomaly samples for each category, resulting in 1,725 images in total. Anomalous samples are further categorized into 15 types of anomaly, each annotated with pixel-level ground truth. The VisA dataset consists of 12 object categories and includes 9,621 normal and 1,200 anomalous images. For a fair comparison, we adopt the reorganized VisA dataset following the RealNet framework \citep{realnet}, which provides a predefined training/test split and restructures VisA to align with MVTec AD's category and file hierarchy.

\textbf{Metrics.} Following previous works \citep{realnet,pang2025context}, we evaluate image-level anomaly detection using the Area Under the Receiver Operating Characteristic Curve (AUROC), which measures the model's ability to distinguish between normal and anomalous images across varying decision thresholds. For pixel-level anomaly localization, we employ both Pixel AUROC and Per-Region Overlap (PRO). Pixel AUROC evaluates the per-pixel classification accuracy, while PRO assesses the overlap between predicted and ground-truth anomalous regions. These metrics are widely adopted in anomaly detection research and provide complementary perspectives on detection performance.

\subsection{Implementation details}
Our WDAM module operates as a plug-and-play component for existing anomaly detection methods. We evaluate its effectiveness by integrating it with three established baselines: CutPaste \citep{cutpaste}, DRAEM \citep{draem}, and PatchCore \citep{patchcore}.

\textbf{CutPaste}. CutPaste utilizes WideResNet50-2 as its backbone network. While the original CutPaste localizes anomalies using GradCAM-generated heatmaps, we leverage feature maps from the fourth layer of the backbone, which consistently yields better performance empirically. While the original CutPaste method uses a three-class classification task (normal, cutpaste scar, cutpaste), we replace the cutpaste class with our synthetic samples and retain the other two classes. Additionally, we insert our WDAM into each bottleneck block in layer1 of the backbone.

Our optimization process comprises two distinct training stages. The first stage replicates the CutPaste methodology, employing stochastic gradient descent (SGD) with an initial learning rate of 0.03 across 500 epochs. In the second stage, we integrate both our synthetic anomalies and WDAM, switching to the Adam optimizer with an initial learning rate of 0.01 that is subsequently reduced to 0.001 upon WDAM incorporation.

\textbf{DRAEM}. We use the official DRAEM implementation with the original hyperparameters, adding a validation phase. WDAM modules are integrated after the first three blocks in DRAEM's five-block architecture.

\textbf{PatchCore}.
We utilize the official PatchCore implementation with its original hyperparameters. PatchCore typically employs two experimental settings: (1) a baseline using WideResNet50-2 with 224×224 input, and (2) an ensemble approach combining WideResNet101, ResNeXt101, and DenseNet201 backbones with 320×320 input. In our implementation, we adopt a modified ensemble strategy that combines results from both the standard WideResNet50-2 and its fine-tuned version with our WDAM, using an input size of 256×256.


\textbf{Stable Diffusion}.
We utilize the Stable Diffusion WebUI repository for anomaly synthesis. The main generation parameters on MVTec AD dataset are configured as follows: the sampling method is set to DPM++2M with 20 sampling steps; the classifier-free guidance (CFG) scale is set in the range of 5 to 10; and the denoising strength is set between 0.5 and 0.85.

All experiments were implemented in PyTorch and conducted on a single NVIDIA RTX A6000 GPU. The Haar wavelet was employed as the basis function in all wavelet transform operations.


\subsection{Results and analysis}
\cref{tab:cutpaste-mvtec} presents a comparative evaluation of the original CutPaste method versus our proposed modifications. \textbf{FMAS} refers to the substitution of the original cutpaste class in the three-class classification task with synthetic anomalies generated by our synthesized strategy. In CutPaste, the scar class generates anomalies randomly across the entire image, which may cause label inconsistency when synthesized anomalies appear on the background rather than the object. \textbf{Foreground Mask (FM)} resolves this issue by constraining CutPaste's scar anomalies to object regions only. Our final model combines FMAS, Foreground Mask, and WDAM. All results represent averages across three independent runs with different random seeds.

From the results, we observe that the progressive integration of our modifications leads to consistent improvements in overall performance. From the baseline CutPaste to our final model, the average image-level AUROC increases by 4.77\%, the pixel-level AUROC by 5.94\%, and the PRO by 14.53\%. The most significant improvements are observed with +FMAS, which leverages our synthetic data, yielding gains of +3.29\%, +4.18\%, and +10.58\% across the three metrics, respectively. WDAM also contributes performance improvements of +1.46\%, +1.40\%, and +4.35\%. FM yields notable improvements in the pill, screw and toothbrush categories, as the objects in these classes occupy relatively small portions of the image.

The quantitative results of CutPaste and our modifications on the VisA dataset are shown in \cref{tab:cutpaste-visa}. Following \citep{EawT}, we report only the image-level AUROC and pixel-level AUROC. All results represent averages across three independent runs with different random seeds. Similar to the results on the MVTec AD dataset, our final model achieves significant improvements of +4.6\% and +1.6\% on these two metrics, respectively. It outperforms \textbf{EawT} \citep{EawT} and \textbf{AST} \citep{AST} in terms of image-level AUROC. These results demonstrate that our method has strong generalizability across different datasets, making it well suited for various industrial scenarios and product types. It is worth noting that, the improvements of using synthetic data generated by our FMAS are evident only after incorporating the foreground mask. This could be attributed to the fact that, compared to MVTec AD, objects in the VisA dataset generally occupy a smaller portion of the image, making the issue of label inconsistency more pronounced.

\begin{table*}[t]
\centering
\caption{Evaluation results of CutPaste, its variants with different modifications, and the other competitors on VisA dataset, with image-level and pixel-level AUROC metrics}
\label{tab:cutpaste-visa}
\renewcommand{\arraystretch}{1.2} 
\setlength{\tabcolsep}{3pt} 

\begin{tabular}{>{\raggedright\arraybackslash}p{2.0cm}|%
                >{\centering\arraybackslash}p{2.2cm}%
                >{\centering\arraybackslash}p{2.2cm}%
                >{\centering\arraybackslash}p{2.2cm}%
                >{\centering\arraybackslash}p{3.2cm}|%
                >{\centering\arraybackslash}p{2.0cm}%
                >{\centering\arraybackslash}p{2.0cm}}
\toprule
\textbf{Category} & \textbf{CutPaste} & \textbf{+FMAS} & \textbf{+FM} & \textbf{+WDAM (Final Model)} & \textbf{EawT} \citep{EawT} & \textbf{AST} \citep{AST}\\
\midrule
candle & (94.7, 93.5) & (93.1, 94.8) & (94.3, \underline{95.8}) & (\textbf{98.0}, 93.4) & (86.8, \textbf{98.6}) & (\underline{97.1}, 93.6) \\
capsules & (84.4, \underline{83.9}) & (72.7, 74.7) & (80.7, 79.0) & (\underline{85.2}, 77.7) & (\textbf{91.2}, \textbf{99.3}) & (81.5, 80.1) \\
cashew & (94.9, 88.3) & (90.2, 90.5) & (\underline{95.3}, \textbf{96.0}) & (\textbf{95.6}, \underline{92.7}) & (88.4, 81.8) & (93.0, 86.3) \\
chewinggum & (97.7, \underline{97.9}) & (97.5, 97.2) & (\underline{99.1}, 93.3) & (99.0, 85.3) & (98, \textbf{98.8}) & (\textbf{99.3}, 87.7) \\
fryum & (93.9, \textbf{97.0}) & (90.4, 88.1) & (95.8, 93.9) & (\underline{97.1}, \underline{95.4}) & (\textbf{97.5}, \underline{95.4}) & (96.8, 79.3) \\
macaroni1 & (85.1, 78.2) & (81.9, 90.2) & (\textbf{97.0}, \underline{91.5}) & (\underline{95.6}, 91.0) & (89.4, \textbf{99.9}) & (75.0, 89.7) \\
macaroni2 & (71.3, 93.1) & (72.9, 95.0) & (86.2, \underline{96.7}) & (\underline{87.9}, 96.0) & (76.4, \textbf{99.4}) & (\textbf{96.9}, 92.9) \\
pcb1 & (95.1, 84.3) & (95.3, 84.1) & (\underline{96.2}, \underline{95.2}) & (95.9, 95.0) & (91.3, \textbf{99.7}) & (\textbf{97.5}, 85.4) \\
pcb2 & (92.3, 94.1) & (88.0, \textbf{97.6}) & (91.7, 95.2) & (\textbf{96.8}, 95.2) & (\underline{95.5}, \underline{97.1}) & (94.4, 83.0) \\
pcb3 & (87.1, \underline{96.7}) & (86.5, 93.2) & (92.6, 91.3) & (93.7 85.8) & (\underline{97.1}, \textbf{99}) & (\textbf{98.4}, 87.3) \\
pcb4 & (97.0, 85.1) & (96.8, 86.4) & (\underline{97.8}, \underline{96.1}) & (\textbf{99.1}, 93.4) & (\textbf{99.1}, \textbf{97.8}) & (97.6, 88.0) \\
pipe fryum & (94.6, 84.0) & (93.6, 95.8) & (\textbf{99.8}, \underline{96.4}) & (\underline{99.6}, 95.1) & (94.9, \textbf{98.6}) & (93.4, 87.4) \\
\midrule
\textbf{Average} & (90.7, 89.7) & (88.2, 90.6) & (\underline{93.9}, \underline{93.4}) & (\textbf{95.3}, 91.3) & (92.1, \textbf{97.1}) & (93.4, 87.4) \\
\bottomrule
\end{tabular}
\end{table*}

\begin{table*}[t]
\centering
\caption{Image-level AUROC and pixel-Level AUROC results on the MVTec AD Dataset. Comparison of DRAEM \citep{draem}, DRAEM with the proposed WDAM, and other methods, including CFLOW-AD \citep{cflow-ad}, MSTUnet \citep{MSTUnet}, MambaAD \citep{mambaad} and DRAEM+SSPCAB \citep{sspcab}}
\label{tab:draem-mvtec}
\renewcommand{\arraystretch}{1.2} 
\setlength{\tabcolsep}{3pt} 


\resizebox{\linewidth}{!}{
\begin{tabular}{>{\raggedright\arraybackslash}p{2.2cm}|%
                >{\centering\arraybackslash}p{2.33cm}%
                >{\centering\arraybackslash}p{2.33cm}%
                >{\centering\arraybackslash}p{2.33cm}|%
                >{\centering\arraybackslash}p{2.33cm}%
                >{\centering\arraybackslash}p{2.33cm}%
                >{\centering\arraybackslash}p{2.33cm}
                }
\toprule
\textbf{Category} & \textbf{CFLOW-AD} \citep{cflow-ad} & \textbf{MSTUnet} \citep{MSTUnet} & \textbf{MambaAD}\citep{mambaad} & \textbf{DRAEM} \citep{draem} & \textbf{+SSPCAB} \citep{sspcab} & \textbf{Final Model} \\
\midrule
Bottle       & (99.0, 96.8)          & (\textbf{100.0}, \textbf{99.0}) & (\textbf{100.0}, \underline{98.8}) & (\underline{99.5}, 98.5) & (98.4, \underline{98.8}) & (99.2, 98.4) \\
Cable        & (\underline{97.6}, 93.5) & (91.4, 89.9)          & (\textbf{98.8}, 95.8)            & (95.3, 94.4)          & (96.9, \underline{96.0})  & (97.4, \textbf{96.5}) \\
Capsule      & (\underline{99.0}, 93.4) & (98.4, \underline{95.7})       & (94.4, \textbf{98.4})            & (96.6, 90.8)          & (\textbf{99.3}, 93.1)  & (97.0, 91.6) \\
Carpet       & (99.3, 97.7)          & (\textbf{99.9}, \underline{98.3}) & (\underline{99.8}, \textbf{99.2}) & (97.9, 95.3)          & (98.2, 95.0)          & (99.6, 95.8) \\
Grid         & (\underline{99.0}, 96.1) & (\textbf{100.0}, \textbf{99.7}) & (\textbf{100.0}, 99.2)            & (\textbf{100.0}, \textbf{99.7}) & (\textbf{100.0}, \underline{99.5})  & (\textbf{100.0}, \textbf{99.7}) \\
Hazelnut     & (\underline{99.0}, 96.7) & (\textbf{100.0}, 99.3)         & (\textbf{100.0}, 99.0)            & (\textbf{100.0}, \underline{99.7}) & (\textbf{100.0}, \textbf{99.8})  & (\textbf{100.0}, 99.6) \\
Leather      & (\underline{99.7}, 99.4) & (\textbf{100.0}, \underline{99.5}) & (\textbf{100.0}, 99.4)            & (\textbf{100.0}, 99.2) & (\textbf{100.0}, 99.5) & (\textbf{100.0}, \textbf{99.6}) \\
Metal nut    & (98.6, 91.7)          & (\textbf{100.0}, \underline{99.3}) & (\underline{99.9}, 96.7)          & (99.8, \textbf{99.6})  & (\textbf{100.0}, 98.9)  & (\textbf{100.0}, \underline{99.3}) \\
Pill         & (\underline{99.0}, 95.4) & (97.4, 97.6)          & (97.0, 97.4)                      & (97.6, \textbf{99.6})  & (\textbf{99.8}, 97.5)  & (97.2, \underline{98.4}) \\
Screw        & (98.9, 95.3)          & (\textbf{100.0}, 97.4)         & (94.7, 99.5)                      & (99.1, \underline{99.6}) & (97.9, \textbf{99.8})  & (\underline{99.4}, \underline{99.6}) \\
Tile         & (98.0, 94.3)          & (\textbf{100.0}, \textbf{99.7}) & (\underline{98.2}, 93.8)          & (\textbf{100.0}, \textbf{99.7}) & (\textbf{100.0}, \underline{99.3}) & (\textbf{100.0}, \textbf{99.7}) \\
Toothbrush   & (\underline{98.9}, 95.1) & (\textbf{100.0}, \underline{99.1}) & (98.3, 99.0)                      & (\textbf{100.0}, \textbf{99.3}) & (\textbf{100.0}, 98.1) & (\textbf{100.0}, \underline{99.1}) \\
Transistor   & (\underline{98.0}, 81.4) & (96.3, 74.6)          & (\textbf{100.0}, \textbf{96.5})    & (95.0, 88.2)          & (92.9, 87.0)          & (95.0, \underline{88.8}) \\
Wood         & (96.7, 95.8)          & (\textbf{100.0}, \textbf{98.0}) & (98.8, 94.4)                      & (\textbf{100.0}, 97.2) & (\underline{99.5}, 96.8) & (\textbf{100.0}, \underline{97.4}) \\
Zipper       & (99.1, 96.6)          & (\textbf{100.0}, 98.9)         & (\underline{99.3}, 98.4)          & (\textbf{100.0}, 98.5) & (\textbf{100.0}, \underline{99.0})  & (\textbf{100.0}, \textbf{99.2}) \\
\midrule
\textbf{Average} & (98.6, 94.6)          & (\underline{98.9}, 96.4)       & (98.6, \textbf{97.7})          & (98.7, 97.3)          & (\underline{98.9}, 97.2)  & (\textbf{99.0}, \underline{97.5}) \\
\bottomrule
\end{tabular}
}
\end{table*}

\begin{table*}[t]
\centering
\caption{Evaluation results of PatchCore, its variants with different modifications, and other methods, including THFR\citep{thfr}, RD++\citep{rd++}, and Dinomaly ViT-B\citep{dinomaly}, on the MVTec AD dataset using image-level AUROC, pixel-level AUROC, and PRO metrics.}
\label{tab:patchcore}
\renewcommand{\arraystretch}{1.2} 
\setlength{\tabcolsep}{3pt} 
\resizebox{\linewidth}{!}{
\begin{tabular}{l|ccc|ccc}               
\toprule
\textbf{Dataset} & \textbf{THFR}\citep{thfr} & \textbf{RD++}\citep{rd++} & \textbf{Dinomaly ViT-B}\citep{dinomaly} & \textbf{PatchCore} & \textbf{PatchCore Ensemble} & \textbf{Final Model} \\
\midrule
MVTec AD & (99.2, 98.2, 95.0) & (99.4, \underline{98.3}, 95.0) & (\textbf{99.6}, \textbf{98.4}, 94.8) & (99.1, 98.1, 97.4) & (\underline{99.5}, 98.2, \underline{97.6}) & (99.0, \underline{98.3}, \textbf{97.7})\\
\bottomrule
\end{tabular}
}
\end{table*}

\begin{figure*}
\centering
\includegraphics[width=\textwidth]{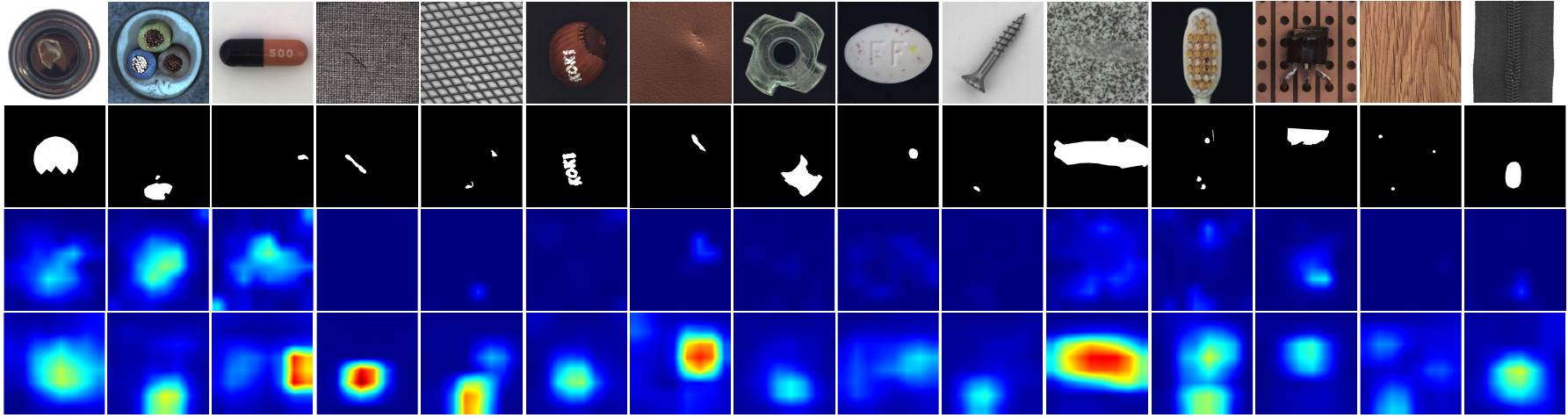}
\vspace{-0.6cm}
\caption{Qualitative results of CutPaste \citep{cutpaste} and our method on MVTec AD dataset. From top to down, original image, ground-truth mask, detection heatmap of CutPaste, and detection heatmap of our method. }
\label{fig:mvtec_heatmap}
\end{figure*}

\begin{figure*}
\centering
\includegraphics[width=\textwidth]{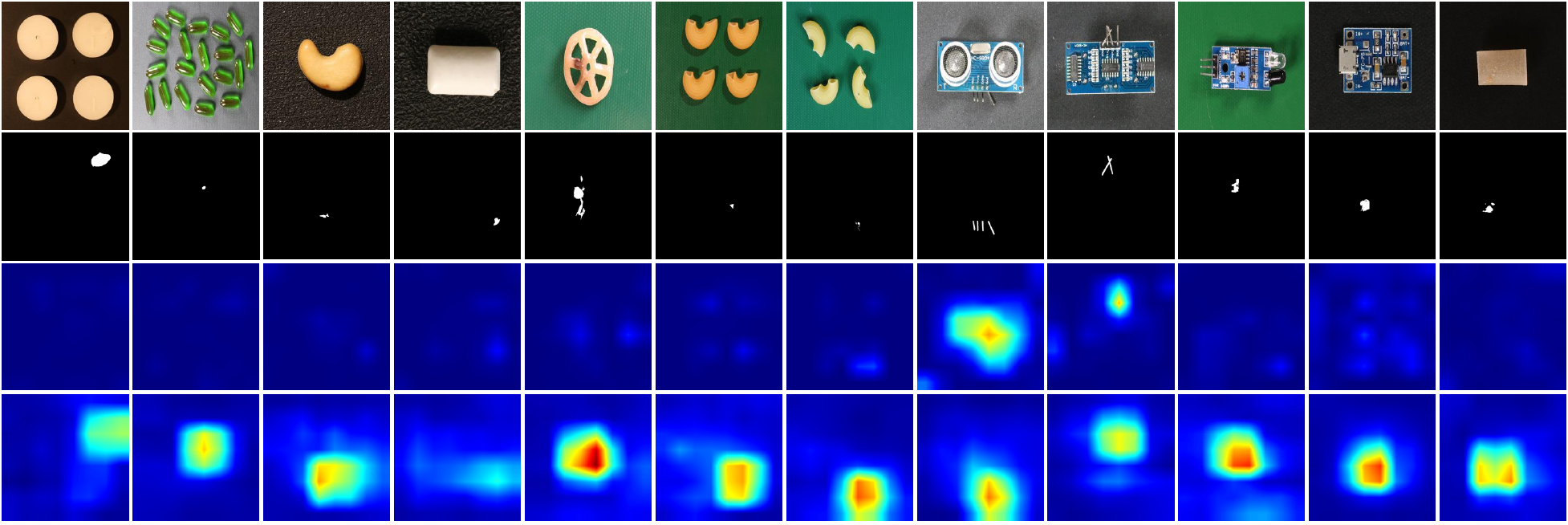}
\vspace{-0.6cm}
\caption{Qualitative results of CutPaste \citep{cutpaste} and our method on VisA dataset. From top to down, original image, ground-truth mask, detection heatmap of CutPaste, and detection heatmap of our method.}
\label{fig:visa_heatmap}
\end{figure*}

\begin{table}[htbp]
\centering
\caption{Module-level ablation results on MVTec AD and VisA dataset, based on the CutPaste.}
\label{tab:component-ablation}
\begin{tabular}{l|cc|c}
\toprule
Dataset & FMAS & WDAM & Metrics \\
\midrule
\multirow{4}{*}{MVTec AD}  &            &       & (93.23, 82.34, 57.51) \\
                           & \checkmark &       & (\underline{96.54}, \underline{86.88}, \underline{67.69}) \\
                           &       & \checkmark & (96.20, 84.91, 66.35) \\
                           & \checkmark & \checkmark & (\textbf{98.00}, \textbf{88.28}, \textbf{72.04}) \\
\midrule
\multirow{4}{*}{VisA}      &            &       & (90.68, 89.68, 65.65) \\
                           & \checkmark &       & (\underline{93.86}, \textbf{93.37}, \textbf{75.82}) \\
                           &       & \checkmark & (90.21, \underline{91.74}, 71.42) \\
                           & \checkmark & \checkmark & (\textbf{95.29}, 91.33, \underline{72.22}) \\
\bottomrule
\end{tabular}
\end{table}

\begin{table}[!t]
\centering
\caption{Ablation analysis of WDAM integration within CutPaste on the MVTec AD dataset.}
\label{tab:layer-ablation}
\begin{tabular}{p{0.8cm}p{0.8cm}p{0.8cm}p{0.8cm}p{3cm}}
\toprule
Layer1 & Layer2 & Layer3 & Layer4 & Metrics \\
\midrule
  &           &           &           & (96.54, 86.88, 67.69) \\
\checkmark &           &           &           & (\textbf{98.00}, \textbf{88.28}, \underline{72.04}) \\
\checkmark & \checkmark &           &           & (\underline{97.92}, \underline{87.99}, 71.70) \\
\checkmark & \checkmark & \checkmark &           & (97.39, 87.92, \textbf{72.09})      \\
\checkmark & \checkmark & \checkmark & \checkmark & (97.45, 87.70, 71.60) \\
\bottomrule
\end{tabular}
\end{table}

\begin{table}[htbp]
\centering
\caption{Comparison of WideResNet50-2 and WideResNet50-2+WDAM in terms of model complexity, efficiency, and accuracy.}
\renewcommand{\arraystretch}{1.2}
\begin{tabular}{lcc}
\hline
Metrics & WideResNet50-2 & +WDAM \\
\hline
Param. (MB)        & 67.40 & 67.59 \\
FLOPs (G)          & 29.92 & 29.94 \\
Inference Time (ms) & 5.71 & 7.35 \\
I-AUROC (\%)       & 96.54 & 98.00 \\
\hline
\end{tabular}
\label{tab:efficiency} 
\end{table}

\begin{table*}
\caption{Average wavelet sub-bands weights learned by the WDAM on the MVTec AD dataset. Each sample is assigned three groups of weights (LL, LH, HL, HH), representing the relative contributions of the four wavelet sub-bands to anomaly detection. For each class, the three comma-separated values correspond to the three WDAM modules integrated sequentially into three bottleneck layers, as illustrated in Figure~\ref{fig:cutpaste_with_wdam}. Values are averaged across all samples within each module.}\label{tab:learned-wave-weights}
\resizebox{\linewidth}{!}{
\begin{tabular}{lccccc ccccc ccccc}
\toprule
\textbf{Class} & bottle & cable & capsule  & carpet & grid   & hazelnut & leather & metal nut & pill & screw   & tile & toothbrush & transistor & wood  & zipper\\
\midrule
\textbf{LL} & 0,0,1 & 0,1,0 & 0,1,1 & 0,1,0 & 0,1,1 & 0,1,0 & 0,1,0 & 0,1,1 & 0,0,1 & 0,1,0 & 0,1,0 & 0,1,0 & 1,1,1 & 0,1,1 & 0,1,0 \\
\textbf{LH} & 1,1,1 & 1,0,0 & 1,0,1 & 0,0,0 & 1,0,0 & 1,0,1 & 1,0,1 & 1,0,0 & 0,0,1 & 0,0,0 & 1,0,0 & 1,0,1 & 1,0,1 & 1,0,0 & 1,0,0 \\
\textbf{HL} & 1,1,0 & 0,0,0 & 1,1,0 & 0,1,0 & 0,1,0 & 0,0,0 & 1,1,0 & 0,1,0 & 0,1,0 & 0,0,0 & 0,0,1 & 0,0,0 & 1,1,1 & 0,0,0 & 0,0,1 \\
\textbf{HH} & 0,0,0 & 0,0,0 & 1,0,0 & 0,0,1 & 1,0,0 & 0,0,1 & 0,0,0.8 & 0,0,1 & 0,0,0 & 0,1,0 & 1,0,1 & 1,0,1 & 0,0,0 & 0,0,0 & 1,0,0 \\
\bottomrule
\end{tabular}
}
\end{table*}

\cref{tab:draem-mvtec} compares the performance of DRAEM before and after integrating the our WDAM on the MVTec AD dataset, along with five recent competitors.
Among them, SSPCAB, proposed in~\citep{sspcab}, is also a plug-and-play module designed for anomaly detection. It should be noted that SSPCAB and WDAM are independently integrated into the DRAEM framework, and the results reported reflect their individual effects rather than a combination of both. However, our method surpasses it in both image-level AUROC and pixel-level AUROC.
With the incorporation of WDAM, DRAEM achieves the highest 99.0\% image-level AUROC among all listed methods and the second-best pixel-level AUROC of 97.5\%, slightly 0.2\% lower than MambaAD. 

\cref{tab:patchcore} shows the performance of PatchCore and our final model on the MVTec AD dataset. Our model achieves 98.3\% pixel-level AUROC and 97.7\% PRO, outperforming the baseline PatchCore Ensemble on both metrics, even though PatchCore leverages an ensemble of three deep backbones.


\subsection{Further analysis}
To systematically analyze the contributions of different components in our framework, we first conduct an ablation study on the MVTec AD and VisA datasets using CutPaste as the baseline, as summarized in \cref{tab:component-ablation}. FMAS and FM are both data-level designs, and are therefore treated as a single component in this ablation study.
The results indicate that the application of FMAS and WDAM each brings significant improvements over the baseline, and that their joint use achieves the best overall performance.

Building upon this analysis, we further investigate the impact of WDAM when inserted at different positions within the network.

To this end, an ablation study was carried out based on CutPaste, as presented in \cref{tab:layer-ablation}.
In the table, $\checkmark$ denotes the inclusion of WDAM at the corresponding network layer. The first row indicates baseline configuration, where WDAM is excluded from all layers.
The results demonstrate that all four WDAM-enhanced variants outperform the baseline, with the layer1-only insertion emerging as the optimal configuration, achieving peak image-level AUROC and pixel-level AUROC, while maintaining a near-optimal PRO score (97.3\%, second-best overall). Notably, this configuration also exhibits the most favorable computational profile, thus providing the best accuracy-efficiency tradeoff for practical deployment.

\cref{tab:efficiency} compares the changes in model parameters (Param.), floating-point operations (FLOPs), inference time, and image-level AUROC (I-AUROC) of WideResNet50-2 before and after the integration of WDAM. The results indicate that the additional model parameters and computational overhead introduced by WDAM are relatively small.

\subsection{Visualization}

\cref{fig:mvtec_heatmap,fig:visa_heatmap} present layer4 backbone feature visualizations comparing CutPaste \citep{cutpaste} with our method across MVTec AD and VisA datasets, respectively. The results demonstrate that while CutPaste's heatmaps exhibit diffuse anomaly responses (pale yellow/blue regions) with significant background noise (false detections) and weak activations in true anomaly areas, our method generates precisely concentrated heatmaps (intense red clusters) that align closely with ground-truth masks while effectively suppressing background artifacts.

WDAM transforms features into the wavelet domain and assigns different weights to each wavelet sub-band (LL, LH, HL, HH), which are then inverse transformed back to the spatial domain through IDWT to enhance anomaly features. \cref{tab:learned-wave-weights} presents the average wavelet sub-band weights learned by the WDAM module on the MVTec AD dataset. Each column represents the weights learned by one module. The results indicate that the weights assigned by WDAM to the four wavelet sub-bands vary depending on the category and the position of WDAM within the network. Since WDAM uses residual connections, columns with all zeros indicate that WDAM has no effect at that layer, and the input features are passed directly to the output.

\section{Conclusion}
\label{sec:conclusion}

In this paper, we introduce FMAS, a training-free anomaly-generation pipeline, and WDAM, a plug-and-play wavelet-domain attention module for anomaly detection. FMAS synthesizes realistic anomalies without fine-tuning or class-specific training by combining GPT-generated prompts, SAM-based foreground masks, and Stable Diffusion inpainting, filtered with an LPIPS-based selector. WDAM enhances anomaly cues with minimal overhead by reweighting the LL/LH/HL/HH sub-bands and reconstructing features via IDWT. 

This design is motivated by the observation that many industrial defects manifest as local structural and textural disruptions that break multi-scale statistical regularities rather than global appearance changes, which often leads to uneven responses across wavelet sub-bands. By explicitly modeling and reweighting these wavelet-domain discrepancies, WDAM provides a generally applicable mechanism for enhancing anomaly saliency. 

Extensive experiments across multiple benchmarks and architectures show significant, consistent gains, validating the effectiveness of both the synthetic-data pipeline and WDAM.

Despite the encouraging results, some limitations remain. The anomalies synthesized by FMAS are primarily realistic at the visual and statistical levels, while physical authenticity is not explicitly enforced, which may limit applicability in highly specialized industrial scenarios. Incorporating domain-specific constraints to further improve physical plausibility is a potential direction for future work. In addition, although FMAS is training-free, configuring Stable Diffusion still involves manual parameter selection. The Stable Diffusion parameter settings reported in the implementation details are provided as a practical reference for reproducibility.

\section*{Acknowledgment}
This research was supported in part by the National Natural Science Foundation of China under Grant 62576313, Key R\&D  Program of Zhejiang Province 2025C01075, and Zhejiang Provincial Natural Science Foundation of China under Grant LD24F020016.


\bibliographystyle{cas-model2-names}


\bibliography{nn-refs}

\end{document}